\documentclass[11pt]{article}

\usepackage[final]{acl}

\usepackage{times}
\usepackage{latexsym}

\usepackage[T1]{fontenc}

\usepackage[utf8]{inputenc}

\usepackage{microtype}

\usepackage{inconsolata}

\usepackage{graphicx}

\title{LLM Olympiad: Why Model Evaluation Needs a Sealed Exam}

\author{Jan Christian Blaise Cruz \textnormal{and} Alham Fikri Aji\\
  MBZUAI \\
  \texttt{\{jan.cruz,alham.fikri\}@mbzuai.ac.ae}}

\begin{document}
\maketitle

\begin{abstract}
Benchmarks and leaderboards are how NLP most often communicates progress, but in the LLM era they are increasingly easy to misread.
Scores can reflect benchmark-chasing, hidden evaluation choices, or accidental exposure to test content---not just broad capability.
Closed benchmarks delay some of these issues, but reduce transparency and make it harder for the community to learn from results.
We argue for a complementary practice: an Olympiad-style evaluation event where problems are sealed until evaluation,
submissions are frozen in advance, and all entries run through one standardized harness.
After scoring, the full task set and evaluation code are released so results can be reproduced and audited.
This design aims to make strong performance harder to ``manufacture'' and easier to trust.
\end{abstract}

\section{Introduction}

\paragraph{Premise.}

\begin{figure}[t]
    \centering
    \includegraphics[width=\linewidth]{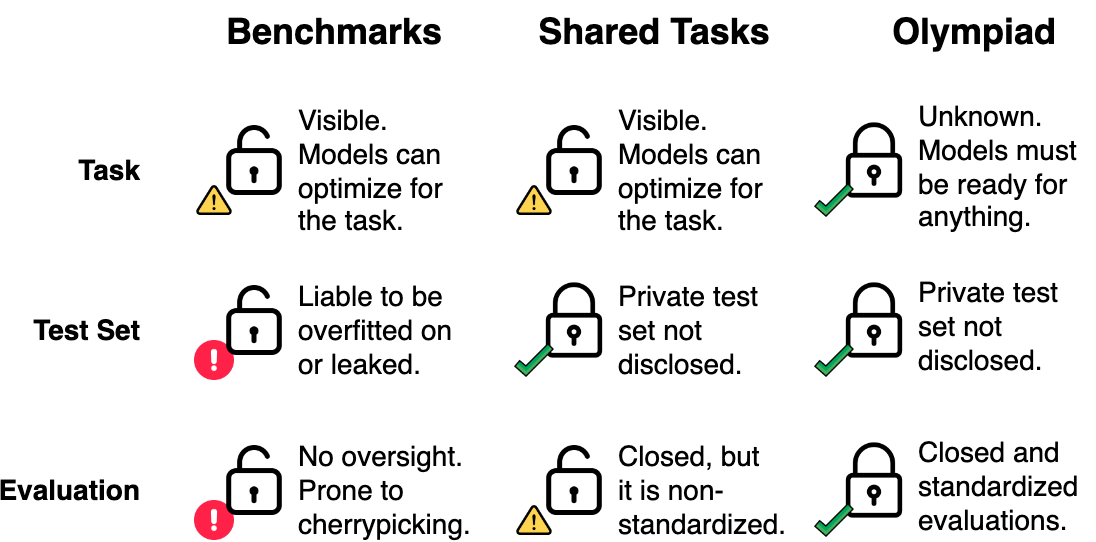}
    \caption{Comparison of evaluation formats across three dimensions. Open benchmarks are fully transparent but expose tasks to optimization and leakage. Shared tasks hide test labels but evaluate against a known target under non-standardized protocols. The proposed Olympiad format seals tasks until evaluation, enforces a standardized harness, and releases all artifacts afterward — combining the strengths of existing formats while mitigating their core weaknesses.}
\label{fig:formats-compare}
\end{figure}

Benchmarks and leaderboards shape how NLP communicates progress.
Yet benchmark outcomes can be \emph{fragile}: small choices in what is measured and how results are aggregated can change which methods look best \citep{dehghani2021benchmark}.
Figure~\ref{fig:formats-compare} previews the three evaluation formats we contrast in this paper.

Two LLM-era forces amplify this fragility.
First, evaluation can become ``leaky'': models trained on web-scale corpora may have already seen test items (or close variants), complicating generalization claims \citep{xu2024benchmark}.
Second, evaluation protocols have many degrees of freedom.
In few-shot settings, even demonstration order can swing performance \citep{lu2022fantastically}, and ``few-shot ability'' can be overstated when prompts and settings are tuned in ways that do not reflect real no-dev evaluation \citep{perez2021true}.

\paragraph{Intuition.}
We take inspiration from Olympiad-style competitions: participants prepare broadly, but the exact problems are revealed only at contest time.
In the International Olympiad such as IMO~\footnote{https://www.imo-official.org}, IOI~\footnote{https://ioinformatics.org}, and other subjects, organizers prepare the problem set in a confidential manner, keeping it sealed until the Olympiad concludes \citep{imo2024shortlist}.
The analogy for LLM evaluation is straightforward: if tasks are known in advance, evaluation naturally becomes a target; if tasks are sealed until scoring, performance is harder to manufacture and easier to interpret.
Secrecy here is \emph{temporary}: tasks and evaluation artifacts are released afterward so the community can audit and reuse them.

\paragraph{Our position.}
We take the position that NLP should complement existing benchmarks with a sealed, Olympiad-style evaluation protocol---an ``LLM Olympiad'' where (i) evaluation problems are sealed until the event, (ii) submissions are frozen and evaluated centrally under a standardized harness, and (iii) all artifacts are released afterward for community audit.
The result is an evaluation where strong performance cannot be manufactured through benchmark-specific optimization or test-set exposure, and where every score comes with a public trail of evidence---tasks, code, and logs---that the community can verify.
The key shift is that the benchmark is not only ``the dataset,'' but the \emph{procedure}: how tasks are solicited, sealed, executed, scored, and then released.

This stance is consistent with prior calls to treat evaluation as infrastructure, including MLPerf's emphasis on shared rules and compliance \citep{reddi2020mlperf} and HELM's push for transparent, standardized reporting \citep{liang2022holistic}.

\paragraph{Roadmap.}
We summarize the limits of today’s benchmarking toolbox in Section \ref{sec:limits}, define the Olympiad-style protocol in Section \ref{sec:olympiad}, describe operational mechanics in Section \ref{sec:mechanics}, and outline a threat model with mitigations and honest limits in Section \ref{sec:threats}.

\section{The Limits of Today's Benchmarking Toolbox}
\label{sec:limits}

Open benchmarks, closed benchmarks, and shared tasks each serve important purposes, and the field is better because they exist.
They also optimize different tradeoffs: open benchmarks prioritize transparency and reuse; closed benchmarks prioritize test secrecy; shared tasks prioritize centralized scoring under a common script.
Our position is not that these formats are misguided, but that in the LLM era they leave gaps for \emph{high-assurance} claims---especially around contamination risk, comparability, and incentives.

\subsection{Open benchmarks}
Open benchmarks are transparent and easy to reproduce, which is a major reason suites like GLUE~\citep{wang2018glue}, MMLU~\citep{hendryckstest2021}, GPQA~\cite{rein2024gpqa}, and other benchmarks became influential.
However, once a benchmark becomes a widely watched target, it invites heavy iteration on prompts and evaluation settings, and small measurement choices can affect perceived winners \citep{dehghani2021benchmark}. For LLMs, this is often amplified by prompt-based evaluation: seemingly minor choices (e.g., demonstration selection or ordering) can shift outcomes, making ``the benchmark score'' less stable than it appears \citep{lu2022fantastically,perez2021true}.

Open benchmarks also face rising contamination risk at web scale \citep{xu2024benchmark}, motivating auditing efforts \citep{li2024open} and deduplication practices \citep{lee2022deduplicating}.
Finally, benchmarks can saturate, prompting cycles like GLUE $\rightarrow$ SuperGLUE \citep{wang2019superglue}.

\subsection{Closed benchmarks}

Closed benchmarks can delay direct test-set overfitting, but trade away community auditability and accessibility, which also affect adaptability by the community.
Without access to tasks and evaluation details, it is harder to interpret why a score changed, or what it teaches the field.
This trust bottleneck motivates calls for transparent, standardized reporting in evaluation \citep{liang2022holistic}.
Closed benchmarks also do not remove incentives to iterate privately and selectively disclose favorable results \citep{singh2025leaderboard}.

Some benchmarks combine both open and closed data access by utilizing two different test sets: one for public release and one for private use~\cite{chollet2025arc, phan2025lastexam}. This approach balances accessibility with privacy; developers can easily measure progress on the public set, then later request an evaluation of their model on the private set. However, this approach still falls short, as practitioners may over-optimize for the specific task.

\subsection{Shared tasks}
Shared tasks improve comparability by running centralized evaluation on hidden test labels.
However, they typically announce the task and format well in advance; strong results can reflect targeted engineering for a known target rather than general preparedness. This is a feature for shared tasks---they are designed to measure adaptation to a specific target---but it differs from the goal of an ``exam-style'' evaluation that probes broad preparedness under surprise problem statements.

For LLM systems, remaining degrees of freedom (prompting conventions, tool use, and compute budgets) can still make comparisons uneven unless the rulebook and harness are strict \citep{reddi2020mlperf,xu2022codabench}. Moreover, many shared-task submissions are highly task-specific. While this is not necessarily wrong, it typically implies that such tasks are not intended for evaluating the capabilities of general-purpose models. Once tasks are released, they can eventually enter training corpora, reintroducing long-term leakage risk \citep{xu2024benchmark}.

It is worth noting that some shared tasks already incorporate elements of the protocol we propose.
SemEval tasks \cite{piskorski2025semeval,muhammad2025semeval} release test data only during the evaluation window; WMT \cite{pakray2025findings,okabe2025findings} has moved toward increasingly controlled evaluation protocols; and the TREC evaluation rounds \cite{arguello2026overview,lawrie2026overview} in information retrieval have long used surprise topics with centralized judging.
We view these as important precedents.
What the Olympiad adds beyond the strongest shared tasks is twofold: first, \emph{task types} are also sealed---participants do not know whether they will face extraction, reasoning, or something else entirely---so models must be broadly prepared rather than task-adapted; second, the harness enforces a single execution and reporting pipeline across all submissions, removing participant-side degrees of freedom that shared tasks typically leave open.

\subsection{Takeaway}
These tools remain valuable, but none consistently provides the combination of (i) sealed evaluation during measurement, (ii) standardized methodology at run time, and (iii) post-hoc auditability after release.
This gap matters more than it used to.
Benchmark rankings increasingly drive high-stakes decisions: which models get deployed, which safety claims are trusted, and where research funding flows.
When those rankings rest on evaluation procedures that are fragile, potentially contaminated, or difficult to reproduce, the downstream consequences are real---and growing.
This motivates an Olympiad-style complement that intentionally borrows strengths from each format: secrecy \emph{during} evaluation (to reduce targeting), centralized execution (to reduce protocol cherry-picking), and transparency \emph{after} evaluation (to enable replication and learning).

\begin{figure*}[t]
    \centering
    \includegraphics[width=\linewidth]{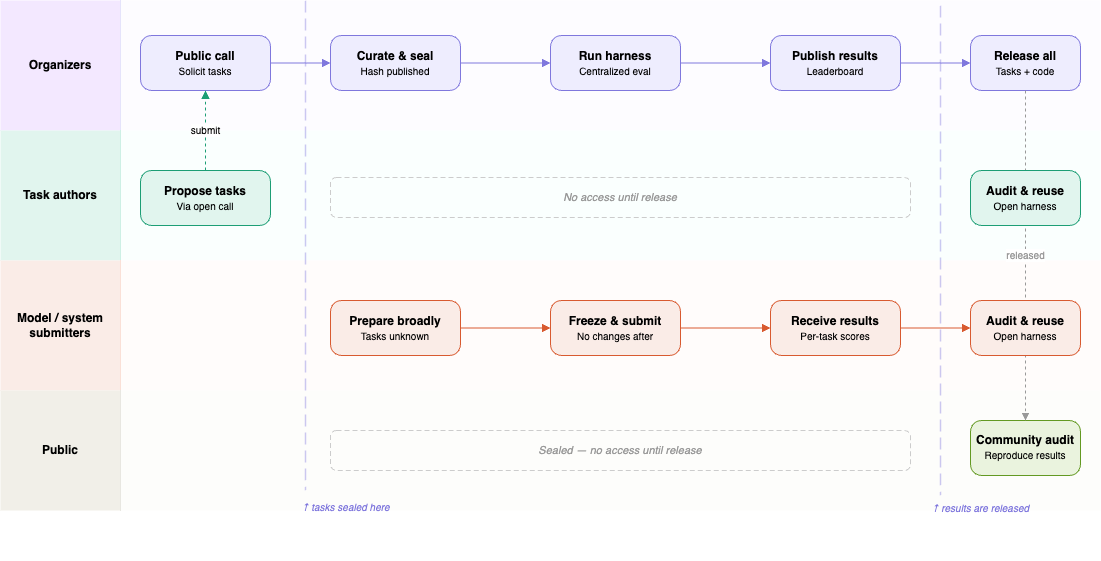}
    \caption{End-to-end flow of the proposed LLM Olympiad. Task authors propose problems via an open call; organizers curate and seal the task bundle before any model submission is evaluated. Model and system submitters prepare broadly without knowledge of task content and freeze their submissions in advance. Evaluation is run centrally under a standardized harness, with scores released to submitters and all artifacts — tasks, scoring code, and harness — released publicly afterward for community audit and reuse.}
\label{fig:olympiad-flow}
\end{figure*}

\section{The LLM Olympiad: A Sealed Evaluation Protocol}
\label{sec:olympiad}

We propose an Olympiad-inspired evaluation event for language models: the process is predictable, but the problems are not revealed until scoring.
The goal is not to replace existing benchmarks, but to add an occasional ``exam'' that supports higher-assurance claims.

\subsection{Expectations}
A reasonable ``LLM Olympiad'' should be predictable in rules even if tasks are a surprise:
\begin{itemize}
  \item \textbf{Rules are public up front:} interface, budgets, scoring, syllabus, and reporting.
  \item \textbf{Problems are sealed:} task content (and even task types) are not public before evaluation.
  \item \textbf{Submissions are frozen:} artifacts or endpoints are committed before scoring.
  \item \textbf{Evaluation is centralized:} one harness, one environment (as far as possible), one reporting pipeline.
  \item \textbf{Artifacts are released after:} tasks and harness are published so results can be reproduced and critiqued.
\end{itemize}
Figure~\ref{fig:olympiad-flow} summarizes the end-to-end workflow.

\subsection{Design principles}
The protocol rests on two separation mechanisms.
First, \textbf{task sealing}: evaluation problems---and even problem types---are kept confidential until scoring begins, so that participants cannot target specific tasks.
Second, \textbf{submission freezing}: model artifacts or endpoint versions are committed before any task content is revealed, preventing last-minute tuning.
When feasible, submissions are also anonymized during scoring to reduce bias in any human-judged components.
Crucially, neither form of separation is permanent: tasks, scoring code, and analysis are released after the event so the community can audit, reproduce, and build on the results \citep{liang2022holistic}.

\subsection{Why it is needed}
Three LLM-era pressures motivate this position, and each is better addressed by an event-based protocol than by incremental repairs to existing formats.

\paragraph{Fragility.}
Benchmark rankings can shift based on seemingly minor evaluation choices---prompt format, demonstration order, aggregation method---making it unclear what a score actually measures \citep{dehghani2021benchmark}.
Existing benchmarks can try to standardize these choices, but in practice each team still runs its own inference pipeline with its own settings.
The Olympiad eliminates this variable entirely: a single harness, controlled by organizers, executes every submission under identical conditions.
Differences in scores reflect differences in models, not differences in evaluation plumbing.
This is not hypothetical: \citet{vendrow2025platinum} constructed GSM8k-Platinum, a cleaned version of the GSM8k benchmark, and found that the ranking of frontier models changed significantly---Claude 3.7 Sonnet and Llama 405B showed identical error counts on the original but diverged sharply on the corrected version, with Llama producing eight times as many errors.

\paragraph{Contamination.}
Models trained on web-scale data may have seen test items or close variants, and this is difficult to detect after the fact \citep{xu2024benchmark,li2024open}.
Refreshing benchmarks periodically helps, but any test set that exists before training begins is potentially exposed.
Sealing tasks until \emph{after} submissions are frozen closes this window: models cannot be tuned on problems that do not yet exist in any accessible form.
Again, this is empirically grounded: when \citet{zhang2024careful} created GSM1k---a fresh set of grade-school math problems designed to mirror GSM8k---they found accuracy drops of up to 13\% for models in the Phi and Mistral families, with systematic overfitting across nearly all model sizes.

\paragraph{Incentive misalignment.}
Leaderboards reward private iteration and selective disclosure---teams can run many configurations and report only the best \citep{singh2025leaderboard}.
This is rational behavior under current rules, but it inflates perceived progress.
\citet{singh2025leaderboard} documented that providers on Chatbot Arena tested up to 27 private model variants before making one public, selecting only the best-scoring version for disclosure.
The Olympiad's freeze-and-commit requirement forces participants to submit a single artifact \emph{before} seeing any tasks, converting the evaluation from a ``best of many tries'' into a one-shot test of general preparedness.

\subsection{Relationship to recent efforts}
The mechanisms above are not novel individually---shared tasks use hidden test sets, MLPerf standardizes execution \citep{reddi2020mlperf}, and Codabench centralizes evaluation platforms \citep{xu2022codabench}.
But as the evidence in the previous section illustrates, each mechanism in isolation leaves a gap that is actively exploited: sealed tasks without a standardized harness still allow protocol cherry-picking; standardized execution without task sealing still permits contamination; and neither prevents selective disclosure without a freeze-and-commit requirement.
The Olympiad's contribution is combining all three in a single event with mandatory post-hoc release.

Several recent efforts address overlapping concerns.
Dynamic benchmarks such as Dynabench \citep{kiela2021dynabench} use adversarial, model-in-the-loop data collection to continuously refresh evaluation sets, while LiveBench \citep{white2024livebench} mitigates contamination by drawing questions from recent information sources on a rolling basis.
Private evaluation services such as Scale AI's SEAL Leaderboards use curated, non-public datasets to resist overfitting.
These are valuable and complementary, but they differ from the Olympiad in important ways.
Continuously refreshed benchmarks reduce contamination risk but do not standardize the execution environment or freeze submissions, so participant-side degrees of freedom remain.
Private evaluations keep test sets sealed but sacrifice the post-hoc transparency that makes results auditable and scientifically reusable.

\subsection{Contest syllabus}
To avoid ``gotcha'' evaluation, the event must publish a contest syllabus in advance:
\begin{itemize}
  \item \textbf{Interface:} allowed inputs/outputs (e.g., text vs.\ structured formats).
  \item \textbf{Budgets:} context/output limits, latency, number of calls, and tool rules (if any).
  \item \textbf{Submission contract:} what counts as ``frozen'' (container/weights vs.\ endpoint commitment).
  \item \textbf{Scoring and reporting:} metrics, aggregation policy, and what will be released after evaluation \citep{liang2022holistic}.
\end{itemize}

To make this concrete, Figure~\ref{fig:syllabus-example} shows an illustrative excerpt of a contest syllabus.
(This is an example for exposition, not a prescription.)

\begin{figure}[t]
\centering
\fbox{
\begin{minipage}{0.95\linewidth}
\footnotesize
\textbf{Example syllabus excerpt (illustrative).}\\[0.5ex]
\textbf{Tracks:} Model (no tools), System (tools allowed via organizer proxy).\\
\textbf{Input/Output:} Input is plain text (or JSON with fields \texttt{id}, \texttt{prompt});
output must be JSON with field \texttt{answer}.\\
\textbf{Budgets:} max input length 12{,}000 characters; max output 1{,}000 characters; max 30s wall-clock per item.\\
\textbf{System tools (if enabled):} retrieval only from organizer-provided corpus; max 3 tool calls per item; all tool calls logged.\\
\textbf{Scoring:} per-task metrics published after evaluation; overall score is macro-average across tasks.\\
\textbf{Release:} after results, tasks + scoring code + harness released; closed endpoints release aggregate logs.
\end{minipage}}
\caption{Illustrative excerpt of a contest syllabus. The event preserves surprise in task content while making rules and constraints predictable.}
\label{fig:syllabus-example}
\end{figure}

In short: \emph{surprise in content, predictability in rules.}

\section{Mechanics: How the LLM Olympiad Would Run}
\label{sec:mechanics}

This section specifies the operational protocol: how tasks enter the event, how they are sealed, how submissions are evaluated, and what is released afterward.
The goal is a process that is predictable, comparable across participants, and auditable after the fact.

\subsection{Soliciting tasks}
The event begins with a public call for ``difficult tasks for LLMs,'' without pre-announcing task families. This process is similar to how international Olympiads collect their problems, which are often submitted by former competitors who remain active in the community. Additionally, these submissions are single-blind, allowing the committee to identify the submitters. Since International Olympiad task contributors are prohibited from training students, task submitters here should likewise be barred from participating in model submissions.

To keep proposals concrete and evaluable, each task submission should include:
\begin{itemize}
  \item a clear problem statement and input/output format;
  \item a scoring plan (preferably automatically verifiable);
  \item a data plan (source, licensing, and what makes it ``fresh'');
  \item a budget estimate (runtime and scale);
  \item a short note on intended failure modes.
\end{itemize}

\paragraph{Task author incentives.}
High-quality tasks are the scarce resource in this proposal, so the event should make task contribution professionally worthwhile.
We recommend explicit recognition for accepted tasks (e.g., a public ``problem setter'' list in the report and workshop materials, a citable task bundle after release, and optional awards such as ``best task'' or ``most diagnostic failure mode'').
To preserve confidentiality, the simplest policy is that task authors do not submit models to the same round; alternatively, a workshop can allow dual participation under strict firewalls (no access to sealed tasks, and clear conflict-of-interest rules).

\subsection{Task Curation}

Once the tasks are submitted, the committee selects which ones will be used for the Olympiad. During this process, the committee may polish or modify the tasks to ensure quality. The task selection process ensures that the final set aligns with the syllabus, maintains a suitable difficulty level, and, most importantly, remains diverse in terms of the task types and capabilities being measured.

Where possible, organizers should perform overlap checks (e.g., screening for duplicate / similarity) to reduce accidental leakage \citep{lee2022deduplicating,xu2024benchmark}.
Organizers should also red-team tasks for ambiguity and scoring loopholes; robustness-oriented evaluation tools provide useful patterns \citep{goel2021robustness}.

\subsection{Sealing problems}
Accepted tasks are bundled into a sealed evaluation set (instances, scoring code, and metadata).
A minimal sealing protocol is:
\begin{itemize}
  \item \textbf{Restricted access:} only a small task committee can access sealed materials, under conflict-of-interest rules.
  \item \textbf{Freeze date:} the task bundle is frozen before any submission is evaluated.
  \item \textbf{Public fingerprint:} organizers publish a hash of an encrypted archive so the community can verify the bundle was not modified post hoc.
\end{itemize}
If a scoring bug is discovered during evaluation, the event should follow a published change-control policy (disclose the bug, patch the harness, and re-run all submissions).
These steps reduce the trust burden and align with an audit mindset \citep{raji2020closing}.

\subsection{Models vs.\ systems (track design)}
To keep results interpretable, we separate \textbf{model} and \textbf{system} submissions.
\begin{itemize}
  \item \textbf{Model track:} the base model plus minimal glue; external tools are disallowed by default.
  \item \textbf{System track:} model + retrieval/tools/orchestration under explicit tool rules, budgets, and logging; this is necessary because system-level failures (e.g., prompt injection) are real \citep{liu2023prompt}.
\end{itemize}
Across both tracks, we support two access modes:
\textbf{open-weights} submissions (containerized artifacts run centrally) and \textbf{closed-weights} submissions (endpoints with version commitments).
Unified execution environments and rulebooks are standard ways to improve comparability \citep{xu2022codabench,reddi2020mlperf}.

\subsection{Evaluation mechanics}
Evaluation should be centralized and routine by design: one harness, one reporting pipeline, and as few implicit choices as possible.
A practical harness should:
\begin{itemize}
  \item enforce a submission freeze window and fixed evaluation period;
  \item enforce budgets (time, tokens, tool calls) and validate output formats;
  \item log versions, decoding settings (when applicable), runtime, failures, and (for systems) tool calls;
  \item report per-task and per-track results with an explicit aggregation policy;
  \item optionally include lightweight consistency checks (e.g., duplicated items, small perturbations) \citep{goel2021robustness}.
\end{itemize}

Two small policies prevent many disputes in practice.
First, define a retry policy (e.g., no retries except one retry on transient infrastructure errors) and how timeouts are scored.
Second, define how stochasticity is handled: either require deterministic decoding for leaderboard scoring, or specify a fixed number of runs and an aggregation rule.

If qualitative judging is included, it should be small, blinded, and reported with agreement procedures, since human evaluation is hard to reproduce \citep{belz2023non}.
LLM-based judges can scale diagnostics but should not be treated as the sole authority \citep{wang2024large}.

\subsection{Results release}
We recommend a two-stage release.
First, publish leaderboards (by track) and a brief analysis report with budgets and constraints \citep{liang2022holistic}.
Second, release the task set, scoring scripts, and harness so results can be audited and reused.

This release package is also a key incentive for participation.
Teams receive a centrally produced, citable evaluation result (with per-task breakdowns and documented constraints), rather than an outcome that depends on private evaluation choices.
A pilot can further encourage participation with lightweight recognition (e.g., awards by track, or a ``reproducible submission'' badge for open-weights artifacts that re-run cleanly under the public harness).

\section{Threat Model}
\label{sec:threats}

We assume strong optimization pressure: participants will push hard for performance, ambiguity will be exploited if it exists, and organizers can make mistakes that bias outcomes.
The goal of this threat model is not to presume adversarial intent, but to design a procedure that makes gaming harder and failures easier to detect.

\subsection{What we aim to protect}
Our protocol is designed to protect three properties:
\textbf{validity} (scores reflect capability rather than exposure or loopholes),
\textbf{comparability} (results are measured under the same constraints), and
\textbf{auditability} (the community can later inspect what happened).

\subsection{Threats and mitigations by phase}

\paragraph{Phase 1: Task creation and sealing.}
The main risks here are (i) \emph{contamination or overlap} with web-scale training data, and (ii) \emph{privileged access} or leakage once tasks are sealed.
Even ``fresh'' tasks can overlap via duplicates or near-duplicates \citep{xu2024benchmark,lee2022deduplicating,li2024open}, and once tasks are released they can eventually enter future training corpora \citep{xu2024benchmark}.
Mitigations are procedural: prefer newly curated data and run similarity screening \citep{lee2022deduplicating};
restrict access to a small task committee under conflict-of-interest rules; and publish a public fingerprint (hash of an encrypted archive) so the community can verify the task bundle was not modified after evaluation \citep{raji2020closing}.
Finally, organizers should red-team tasks for ambiguity and scoring loopholes; robustness-oriented evaluation tools provide useful patterns \citep{goel2021robustness}.

\paragraph{Phase 2: Submission integrity and freezing.}
A sealed test does not remove incentives to iterate privately.
Teams can try many variants and submit only the best, which can amplify leaderboard distortions \citep{singh2025leaderboard,dehghani2021benchmark}.
For closed endpoints, there is an additional risk: \emph{endpoint change} or unlogged behavior changes during the evaluation window.
Mitigations include a hard freeze window, separate leaderboard reporting for open-weights versus closed-weights submissions, and endpoint commitments (version identifiers and timeboxed evaluation).
As a lightweight detection tool, the harness can include small \emph{consistency probes} (duplicate items or minor perturbations) to flag instability without changing the main scoring rules (Figure~\ref{fig:consistency-probe}).

\begin{figure}[t]
\centering
\fbox{
\begin{minipage}{0.95\linewidth}
\footnotesize
\textbf{Example: consistency probe (illustrative).}\\
The harness duplicates a small fraction of items (same input, same budgets).\\[0.5ex]
\textbf{Item:} ``Convert \texttt{3/4} to a percentage.''\\
\textbf{Run A:} \texttt{75\%} \hfill \textbf{Run B:} \texttt{0.75\%}\\
\textbf{Flag:} inconsistent; counted in a stability report (separate from the main score).
\end{minipage}}
\caption{A lightweight consistency probe can detect instability or endpoint drift without changing the main scoring rules.}
\label{fig:consistency-probe}
\end{figure}

\paragraph{Phase 3: Evaluation harness and tool use.}
Centralized evaluation reduces participant-side cherry-picking, but increases the importance of harness choices.
Prompting, decoding, failure-handling, and aggregation can materially affect outcomes \citep{perez2021true}, especially in few-shot settings.
Mitigation here is preregistration: fix decoding, failure-handling, aggregation, and reporting policies \emph{before} evaluation, and disclose any bug-fix process that would require re-running all submissions \citep{liang2022holistic}.
For the system track, tool use can bypass constraints unless access is controlled and logged; prompt injection is a concrete system risk \citep{liu2023prompt}.
Mitigations include tool proxies (organizer-controlled gateways), budgets, and logging for tool calls \citep{liu2023prompt}.

\paragraph{Phase 4: Reporting and interpretation.}
Even if evaluation is clean, results can still be misread if reporting collapses heterogeneous tasks into a single number or if aggregation is unclear.
Leaderboard ordering can depend on aggregation choices \citep{tatiana2021not}.
Mitigation is robust reporting: publish per-task and per-track results, make aggregation explicit, and report budgets alongside scores \citep{liang2022holistic}.
If qualitative evaluation is included, it should be treated as diagnostic rather than the primary ranking signal, since human evaluation can be hard to reproduce \citep{belz2023non} and LLM-based judging can introduce systematic bias \citep{wang2024large}.

\subsection{Risks and honesty limits}
Even with strong procedures, an Olympiad-style evaluation cannot promise perfect guarantees.
Instead, it should make the remaining uncertainty \emph{explicit} so results are interpreted appropriately.

\paragraph{Contamination remains possible.}
Sealing tasks reduces pre-event targeting, but it cannot guarantee zero contamination at web scale \citep{xu2024benchmark}.
Overlap can occur through duplication, paraphrase, or indirect inclusion in training mixtures.
For this reason, Olympiad results should be framed as \emph{stronger evidence} of general preparedness than typical public leaderboards, not as a mathematical proof of ``no exposure.''
This is also why post-hoc release matters: once tasks are public, the community can run overlap checks and scrutinize surprising outcomes.

\paragraph{Closed endpoints are lower assurance than centrally run artifacts.}
Closed endpoints enable broader participation, but they are intrinsically harder to audit.
Endpoints can drift, route requests through undisclosed components, or change behavior under load---sometimes unintentionally.
We therefore recommend labeling open-weights (organizer-run) and closed-weights (endpoint-run) as different assurance tiers in reporting \citep{reddi2020mlperf,xu2022codabench}.
This is not meant to exclude closed models; it is meant to prevent apples-to-oranges comparisons from being overstated.
We acknowledge that version commitments and time-boxing are procedural safeguards, not technical guarantees---they cannot detect hidden routing or undisclosed tool use behind an endpoint.
For this reason, the Olympiad's strongest assurance claims apply to the open-weights track, where organizers control the full execution environment.

\paragraph{The harness can be wrong.}
Centralized evaluation reduces participant-side degrees of freedom, but it concentrates responsibility in the harness.
Bugs, edge cases, and seemingly small choices (timeouts, retries, aggregation) can affect outcomes.
The best defense is procedural: preregister harness policies, require that bug fixes trigger re-running all submissions, and release tasks and code after scoring so the community can replicate and critique the evaluation \citep{liang2022holistic}.

\paragraph{A small task set may not be representative.}
With only a handful of sealed tasks, the Olympiad risks moving the ``benchmark lottery'' \citep{dehghani2021benchmark} upstream: rankings may reflect which capabilities the task committee happened to include rather than broad model quality.
Several design choices can mitigate this.
First, open solicitation (rather than committee-designed tasks) diversifies the pool beyond organizer taste.
Second, the curation stage can enforce explicit coverage targets---requiring tasks from distinct capability families (e.g., extraction, reasoning, generation, robustness) so no single skill dominates the aggregate score.
Third, per-task breakdowns, which we recommend as the primary reporting format, allow the community to assess whether a model's ranking is driven by genuine breadth or by a few favorable tasks.
Even so, a small event will inevitably sample capability space sparsely, and aggregate rankings should be interpreted as one snapshot rather than a comprehensive verdict.
Increasing task diversity across iterations is one of the goals of the pilot plan (Appendix~\ref{sec:appendix-pilot}).

\section{Conclusion}
\label{sec:conclusion}

The LLM era has made evaluation both more important and harder to trust.
Open benchmarks maximize transparency but are easy to target; closed benchmarks protect test sets but reduce scrutiny;
shared tasks improve fairness but often measure performance on a known target.
If we want scores that carry more weight---scores that are harder to game and easier to interpret---we need an additional tool.

We propose an Olympiad-style, sealed evaluation event as that tool.
Seal the problems, freeze submissions, run everyone under one harness, and then release everything afterward so the community can audit
and learn from the outcomes.
This does not replace existing benchmarks; it adds an occasional ``exam'' whose results are more credible when the stakes are high.
If the community adopts even one such event per year, we gain something no current format provides: a shared, reproducible, high-assurance checkpoint that the field can point to and say, ``this is what these models could actually do, measured fairly, on problems no one had seen.''

\section*{Limitations}
This is a position paper: we propose a protocol and do not report results from an implemented Olympiad.
The empirical evidence we cite (Section 3.3) demonstrates that the problems motivating our proposal---contamination, rank instability, and selective disclosure---are real, but it does not constitute a test of the protocol itself; that requires a pilot.
Several operational choices (governance, infrastructure, budgets, and track definitions) will depend on venue constraints and will likely require iteration.
The proposal also introduces a nontrivial governance burden---sealed evaluation requires trusted organizers, conflict-of-interest policies, and change-control procedures that are themselves subject to failure.
Recruiting high-quality task contributions to an unproven event is a further practical challenge, since task authors bear effort and confidentiality constraints without guaranteed return.
Sealed tasks reduce benchmark-chasing but cannot eliminate contamination risk, especially for closed models and web-scale training.
We therefore frame this as a complementary, higher-assurance evaluation layer rather than a replacement for existing benchmarks.

\section*{Ethical Considerations}
This paper proposes an evaluation \emph{procedure} rather than a new model or dataset.
Even so, procedure design has ethical implications because it shapes incentives, participation, and what kinds of claims the community treats as credible.

\paragraph{Access and fairness.}
A sealed-task event could unintentionally privilege well-resourced teams if participation requires large compute budgets or specialized infrastructure.
To mitigate this, the event should publish clear budget classes (time/token/tool limits), provide separate tracks for open-weights and closed endpoints, and report budgets next to scores.
Where feasible, organizers can further improve accessibility by subsidizing evaluation for a small number of academic teams or by offering a low-budget baseline track.

\paragraph{Data provenance and privacy.}
Task proposals should include licensing and provenance information, and organizers should avoid tasks that require personal data or sensitive content.
When tasks involve real-world text, organizers should prefer sources with clear permissions and remove personally identifying information.
When the task set is released after the event, accompanying documentation should describe data sources, intended use, and known limitations.

\paragraph{Dual use and security.}
Some system-oriented evaluations (e.g., prompt injection stress tests) can reveal vulnerabilities that are also useful to attackers.
Organizers should therefore consider responsible release practices for security-sensitive tasks, such as delayed release, partial redaction of exploit strings, or coordinated disclosure when appropriate.
More generally, the analysis report should avoid providing step-by-step guidance for misuse while still reporting scientific findings.

\paragraph{Responsible communication.}
Because competitive leaderboards can encourage overinterpretation, the event should emphasize per-task breakdowns, capability profiles, and documented constraints rather than a single ``winner-takes-all'' number.
Clear labeling of assurance tiers (e.g., open-weights vs.\ closed endpoint) helps prevent results from being presented with more certainty than the protocol can support.

\paragraph{AI Use Disclaimer.}
We acknowledge the use of ChatGPT 5.4 Pro to aid in the proofreading of the final draft of this paper.


\bibliography{custom}

\appendix

\section{Practical Considerations}
\label{sec:appendix-practical}
This appendix summarizes practical design choices for infrastructure, governance, and reporting that affect cost and auditability.
The goal is not to prescribe a single design, but to make the tradeoffs explicit so a workshop can choose a realistic first version.

\subsection{Infrastructure}
The Olympiad needs infrastructure that supports two goals that can be in tension: standardization (everyone is tested the same way)
and accessibility (not every strong model can be submitted as open weights).
A practical setup supports two submission modes.

\paragraph{Open-weights (organizer-run).}
Participants submit a packaged artifact (e.g., a container) that exposes a simple inference interface.
Organizers run it on a fixed hardware class in a controlled environment.
This is the highest-assurance mode because it minimizes uncertainty about what code is executed and under what conditions
\citep{reddi2020mlperf,xu2022codabench}.
To reduce friction, keep the interface stable (e.g., ``given an input file, write an output file,'' or a local server with a fixed schema),
and publish hardware ``classes'' early (even if a pilot supports only one tier).

\paragraph{Closed-weights (endpoint-run).}
Participants provide an endpoint plus a version commitment (and a freeze window).
Organizers run a single evaluation client that enforces rate limits, request budgets, and logging.
These results should be treated as a separate assurance tier in reporting.

Across both modes, several infrastructure choices reduce disputes:
(i) publish a determinism and re-run policy;
(ii) enforce budgets in the harness (max input/output length, max wall-clock time per item, and tool-call limits for systems);
(iii) log enough metadata for post-hoc audit (submission ID, version hash, decoding settings when applicable, runtime, and tool calls);
and (iv) if tools are allowed, route tool access through organizer-controlled proxies so calls can be logged and limited,
which also helps manage known system risks like prompt injection \citep{liu2023prompt}.
Finally, sealed task storage should be treated as core infrastructure: store tasks in a restricted-access repository and freeze archives before evaluation.

\subsection{Governance}
Sealed evaluation increases the need for clear governance.
A workable structure separates roles so no single group has end-to-end control:
a steering committee (rules and release plan), a task committee (curation and sealing),
evaluation operators (running the harness), and a report team (analysis and write-up).
Where possible, task authors should not overlap with evaluation and reporting roles.

Key policies to publish in advance include: a conflict-of-interest policy (submitters do not access sealed tasks),
a confidentiality window (who can access tasks and when tasks will be revealed),
and change control (no changes after freeze except a disclosed bug-fix process that triggers re-running all submissions).
A lightweight appeals process for scoring bugs and harness failures is also important, since most disputes arise from edge cases.
Publishing a public fingerprint (hash) of the sealed bundle before evaluation is a simple audit practice that reduces the trust burden
\citep{raji2020closing}.

\subsection{Reporting}
Reporting is where the Olympiad becomes useful science rather than a one-off contest.
We recommend three layers.

\paragraph{Leaderboards.}
Separate leaderboards by track (model vs.\ system) and by access mode (open-weights vs.\ closed-weights).
Report an overall score alongside per-task breakdowns, and report budgets next to scores, following the spirit of transparent, multi-metric evaluation \citep{liang2022holistic}.
For clarity, a pilot leaderboard can include the following columns:
\emph{submission ID}, \emph{track}, \emph{access mode}, \emph{overall score}, \emph{per-task scores}, \emph{budget class},
\emph{runtime (median)}, and \emph{error rate/timeouts}.

\paragraph{Analysis report.}
Include a short narrative of common failure patterns, a small number of concrete examples (with careful redaction if needed),
and basic stability checks (e.g., duplicated items or small perturbations).
For system submissions, report tool-use statistics (calls and failures) and note any policy violations detected by the harness.

\paragraph{Reproducibility bundle.}
After results are released, publish the sealed tasks, scoring scripts, and evaluation harness.
Release a run manifest that makes the evaluation replayable: submission IDs, versions (or endpoint commitments), budget class,
decoding policy, failure counts, and any exclusions with reasons.
If detailed logs cannot be released (e.g., proprietary endpoints), release aggregate metadata (runtime distribution, error counts, timeout counts)
to support sanity checks.

\begin{figure}[t]
\centering
\fbox{
\begin{minipage}{0.95\linewidth}
\footnotesize
\textbf{Example: run manifest excerpt (illustrative).}\\[0.5ex]
\texttt{submission\_id: S042 \ \ track: model \ \ mode: open-weights}\\
\texttt{artifact\_hash: 9f3c... \ \ budget\_class: A (30s/item)}\\
\texttt{decoding: greedy \ \ max\_output: 1000 chars}\\
\texttt{timeouts: 3/1200 \ \ format\_errors: 1/1200 \ \ notes: none}\\
\end{minipage}}
\caption{Illustrative run manifest fields. Even minimal manifests improve auditability and reduce ambiguity about evaluation conditions.}
\label{fig:run-manifest}
\end{figure}

\section{Expected Submissions}
\label{sec:appendix-submissions}
This appendix gives concrete examples of what we mean by ``task submissions'' and ``model/system submissions.''
Because the Olympiad is designed to keep task content private until evaluation, these examples are illustrative only:
they are meant to clarify format and expectations, not to prescribe the types of tasks that will appear.

\subsection{Task submissions (problem proposals)}
A task submission should be short, concrete, and scorable.
As a rule of thumb, task proposals should fit in 1--2 pages and include:
\begin{itemize}
  \item \textbf{Task summary:} what the model is asked to do, in one paragraph.
  \item \textbf{Input/output contract:} what inputs look like and what outputs must look like (including any structured schema).
  \item \textbf{Scoring:} how correctness is measured; prefer automatic scoring when possible.
  \item \textbf{Data plan:} where instances come from, what makes them ``fresh,'' and licensing/permissions.
  \item \textbf{Budgets:} expected runtime per item and recommended evaluation set size.
  \item \textbf{Failure modes:} what the task is intended to reveal (common model mistakes).
\end{itemize}

\begin{figure}[t]
\centering
\fbox{
\begin{minipage}{0.95\linewidth}
\footnotesize
\textbf{Example task proposal A (illustrative): Structured extraction.}\\
\textbf{Summary:} Given short incident reports, extract fields into a fixed JSON schema.\\
\textbf{Input:} a paragraph of text. \quad
\textbf{Output:} JSON with fields \texttt{category}, \texttt{location}, \texttt{time}, \texttt{entities}.\\
\textbf{Scoring:} schema validity + exact match on normalized fields; partial credit for entity lists.\\
\textbf{Data plan:} newly written synthetic reports with controlled ambiguity; non-overlapping templates.\\
\textbf{Budgets:} 1{,}000 items; 15s/item cap; max output 1{,}000 chars.\\
\textbf{Failure modes:} hallucinated fields, schema breakage, brittle formatting.
\end{minipage}}
\caption{Illustrative task proposal emphasizing verifiable scoring and clear I/O contracts.}
\label{fig:task-proposal-structured}
\end{figure}

\begin{figure}[t]
\centering
\fbox{
\begin{minipage}{0.95\linewidth}
\footnotesize
\textbf{Example task proposal B (illustrative): Evidence-based QA with abstention.}\\
\textbf{Summary:} Answer a question using only the provided evidence, or abstain if unsupported.\\
\textbf{Input:} question + evidence passages. \quad
\textbf{Output:} answer + \texttt{support\_span} (or \texttt{ABSTAIN}).\\
\textbf{Scoring:} answer correctness + penalty for unsupported answers; bonus for correct abstention.\\
\textbf{Data plan:} curated passages created for the event; ensure questions are not quotable web facts.\\
\textbf{Budgets:} 500 items; 30s/item cap; max output 800 chars.\\
\textbf{Failure modes:} confident guessing, citing irrelevant spans, ignoring abstention option.
\end{minipage}}
\caption{Illustrative task proposal where correctness depends on grounding and calibrated refusal.}
\label{fig:task-proposal-grounded}
\end{figure}

\begin{figure}[t]
\centering
\fbox{
\begin{minipage}{0.95\linewidth}
\footnotesize
\textbf{Example task proposal C (illustrative): Stability under innocuous variation.}\\
\textbf{Summary:} The same underlying instruction is presented with mild paraphrases and distractors.\\
\textbf{Input:} instruction variants. \quad
\textbf{Output:} a structured answer (e.g., JSON with \texttt{decision} and \texttt{rationale}).\\
\textbf{Scoring:} correctness on each item + a stability score measuring variance across variants.\\
\textbf{Data plan:} generator produces paraphrases and distractors; instances are sealed until evaluation.\\
\textbf{Budgets:} 300 ``families'' of 3 variants each; 20s/item cap.\\
\textbf{Failure modes:} sensitivity to phrasing, instruction hijacking by distractors.
\end{minipage}}
\caption{Illustrative task proposal where the goal is not only accuracy but consistent behavior.}
\label{fig:task-proposal-stability}
\end{figure}

\subsection{Model and system submissions}
To reduce friction, we recommend that each submission include a short \textbf{submission card} (metadata) plus a runnable artifact (open-weights) or endpoint commitment (closed-weights).
A submission card helps the organizers run the harness and helps readers interpret results later.
A minimal submission card can include:
\begin{itemize}
  \item \textbf{Track and mode:} model vs.\ system; open-weights vs.\ closed-weights.
  \item \textbf{Interface:} how to call the model/system (CLI schema or endpoint schema).
  \item \textbf{Budgets:} maximum context size, maximum output size, and any latency limits.
  \item \textbf{Decoding policy:} greedy vs.\ sampling; if stochastic, how randomness is controlled.
  \item \textbf{Tool policy (systems only):} what tools are used and how calls are logged (if allowed).
\end{itemize}

\begin{figure}[t]
\centering
\fbox{
\begin{minipage}{0.95\linewidth}
\footnotesize
\textbf{Example submission A (illustrative): Open-weights model, organizer-run.}\\
\textbf{Track/mode:} Model / open-weights.\\
\textbf{Artifact:} Docker image exposes a CLI: \texttt{run\_model --input input.jsonl --output out.jsonl}.\\
\textbf{I/O:} each line has \texttt{id} and \texttt{prompt}; output lines must contain \texttt{id} and \texttt{answer}.\\
\textbf{Budgets:} max input 12k chars; max output 1k chars; 30s/item.\\
\textbf{Decoding:} greedy (deterministic).\\
\textbf{Notes:} no external calls; all inference local.
\end{minipage}}
\caption{Illustrative open-weights submission emphasizing a simple, reproducible interface.}
\label{fig:submission-openweights-model}
\end{figure}

\begin{figure}[t]
\centering
\fbox{
\begin{minipage}{0.95\linewidth}
\footnotesize
\textbf{Example submission B (illustrative): Closed-weights endpoint, organizer-run client.}\\
\textbf{Track/mode:} Model / closed-weights.\\
\textbf{Endpoint:} \texttt{POST /infer} with JSON \{\texttt{id}, \texttt{prompt}\} $\rightarrow$ \{\texttt{id}, \texttt{answer}\}.\\
\textbf{Commitment:} version identifier + freeze window agreement (no updates during evaluation).\\
\textbf{Budgets:} declared max context/output; rate limit provided so organizers can schedule runs.\\
\textbf{Decoding:} declared default settings; if stochastic, fixed seed policy or fixed-\#runs policy.
\end{minipage}}
\caption{Illustrative closed-weights endpoint submission. Results should be reported as a separate assurance tier.}
\label{fig:submission-closedweights-model}
\end{figure}

\begin{figure}[t]
\centering
\fbox{
\begin{minipage}{0.95\linewidth}
\footnotesize
\textbf{Example submission C (illustrative): System with retrieval via organizer proxy.}\\
\textbf{Track/mode:} System / open-weights.\\
\textbf{Artifact:} container runs an orchestrator (model + retrieval calls).\\
\textbf{Tool policy:} retrieval only through organizer proxy \texttt{retrieve(query)}; max 3 calls/item; all calls logged.\\
\textbf{Budgets:} 30s/item total; retrieval responses count toward context budget.\\
\textbf{Failure handling:} tool errors must be handled gracefully; tool calls after budget are disallowed.
\end{minipage}}
\caption{Illustrative system submission that allows tools while keeping tool access auditable and budgeted.}
\label{fig:submission-system-rag}
\end{figure}

\begin{figure}[t]
\centering
\fbox{
\begin{minipage}{0.95\linewidth}
\footnotesize
\textbf{Example submission D (illustrative): System with structured outputs.}\\
\textbf{Track/mode:} System / open-weights.\\
\textbf{I/O:} outputs must be valid JSON under a published schema (e.g., \texttt{decision}, \texttt{evidence}, \texttt{confidence}).\\
\textbf{Scoring compatibility:} organizer harness validates schema; invalid JSON is scored as incorrect for that item.\\
\textbf{Benefit:} reduces ambiguity and makes automatic scoring easier for some tasks.
\end{minipage}}
\caption{Illustrative system submission optimized for schema-valid, automatically scorable outputs.}
\label{fig:submission-system-structured}
\end{figure}

\section{Pilot Plan}
\label{sec:appendix-pilot}
This section outlines a minimal pilot suitable for a first workshop iteration.
The aim is to start small, demonstrate value, and iterate.

\subsection{Scope and tracks}
A reasonable pilot can start with two tracks:
(i) \textbf{Model / open-weights} (organizer-run), and
(ii) \textbf{Model / closed-weights} (endpoint-run), reported as a separate assurance tier.
If there is capacity, add a system track as non-competitive in year one (reported, but not ranked against the model track),
so tool governance and harness rules can mature before they affect primary rankings.

\subsection{Timeline}
One workable schedule is:
(1) call for tasks (4--6 weeks);
(2) curation and sealing (4 weeks);
(3) call for submissions (4--6 weeks);
(4) a hard freeze deadline;
(5) evaluation window (1--2 weeks);
(6) results release aligned with the workshop; and
(7) post-hoc release of tasks and code within 2--4 weeks.

\subsection{Submission contract}
Keep submission contracts simple.
Open-weights submissions provide a container exposing a single inference entry point (CLI or local server),
declared context/output limits, and a required ``dry run'' format check.
Closed-weights submissions provide an endpoint plus authentication method (if needed),
a version commitment statement and freeze window agreement, and declared limits so organizers can budget.
In both tracks, require a short metadata form (kept private until reveal) describing decoding settings and any refusal/safety policies that may affect outputs.

See Appendix~\ref{sec:appendix-submissions} for concrete examples of expected task proposals and model/system submissions.

\subsection{Evaluation scale}
A pilot should prioritize finishing cleanly over being large.
A rough target is 5--10 tasks with 100--500 instances each (1k--5k total instances),
plus a small number of consistency probes (duplicates or minor perturbations) to measure stability at low additional cost.
Cap per-instance runtime so the full evaluation can complete in days rather than weeks.

\subsection{Success criteria}
A pilot succeeds if it demonstrates:
(i) operational feasibility (the harness runs reliably and policies reduce disputes);
(ii) comparability (results depend less on hidden evaluation choices);
(iii) learning value (the report surfaces actionable failure patterns); and
(iv) auditability (after release, at least one independent group can reproduce results for a subset of submissions or re-run the harness on a new model).

\subsection{Iteration after year one}
If the pilot is successful, iteration should focus on:
adding a mature system track with stronger tool governance,
expanding task diversity while preserving verifiable scoring,
and maintaining a small unreleased item bank so each year stays fresh without reusing the same problems.


\end{document}